\documentclass{bmvc2k}
\usepackage{multirow}
\usepackage{soul}
\usepackage{booktabs}

\title{ALFred: An Active Learning Framework for Real-world Semi-supervised Anomaly Detection with Adaptive Thresholds}

\addauthor{Shanle Yao}{syao@charlotte.edu}{1}
\addauthor{Ghazal Alinezhad~Noghre}{galinezh@charlotte.edu}{1}
\addauthor{Armin Danesh~Pazho}{adaneshp@charlotte.edu}{1}
\addauthor{Hamed Tabkhi}{htabkhiv@charlotte.edu}{3}

\addinstitution{
 Department of Electrical and Computer Engineering,
 University of North Carolina at Charlotte,
 Charlotte, USA
}

\runninghead{Student, Prof}{BMVC Author Guidelines}

\begin{document}

\maketitle

\begin{abstract}
Video Anomaly Detection (VAD) can play a key role in spotting unusual activities in video footage. VAD is difficult to use in real-world settings due to the dynamic nature of human actions, environmental variations, and domain shifts. Traditional evaluation metrics often prove inadequate for such scenarios, as they rely on static assumptions and fall short of identifying a threshold that distinguishes normal from anomalous behavior in dynamic settings. To address this, we introduce an active learning framework tailored for VAD, designed for adapting to the ever-changing real-world conditions. Our approach leverages active learning to continuously select the most informative data points for labeling, thereby enhancing model adaptability. A critical innovation is the incorporation of a human-in-the-loop mechanism, which enables the identification of actual normal and anomalous instances from pseudo-labeling results generated by AI. This collected data allows the framework to define an adaptive threshold tailored to different environments, ensuring that the system remains effective as the definition of 'normal' shifts across various settings. Implemented within a lab-based framework that simulates real-world conditions, our approach allows rigorous testing and refinement of VAD algorithms with a new metric. Experimental results show that our method achieves an EBI (Error Balance Index) of 68.91 for Q3 in real-world simulated scenarios, demonstrating its practical effectiveness and significantly enhancing the applicability of VAD in dynamic environments.
\end{abstract}
    
\section{Introduction}
\label{sec:intro}

Video Anomaly Detection (VAD) with semi-supervised learning is a crucial domain of computer vision focused on identifying unusual behavior in video streams, with applications ranging from surveillance to healthcare \cite{Bao_2024vand6}. While traditional pixel-based methods \cite{doshi2021online, zhang2023online, karim2024real} often raise privacy and bias concerns, especially for marginalized groups, pose-based approaches offer a privacy-preserving and more equitable alternative by relying on skeletal motion data. However, deploying these models outside controlled environments remains a major challenge, requiring innovations to ensure real-world robustness.

The real world demands that VAD systems contend with a host of practical difficulties. Anomaly definition is inherently relative and context-dependent. For instance, running in a library would be flagged as unusual, whereas jogging in a park is perfectly normal—necessitating systems to adapt their understanding of "normal" and "anomalous" across diverse environments. Real-time detection imposes tight constraints on inference time, requiring models to process streaming video data quickly for timely alerts, yet the computational demands of algorithms often outstrip available resources. Defining clear anomaly boundaries further complicates matters, as many VAD algorithms produce anomaly scores that are not inherently percentages \cite{markovitz2020gepc, hirschorn2023stgnf, noghre2024tsgad}, complicating the establishment of clear decision boundaries. Moreover, standard metrics like AUC-ROC and AUC-PR rely on labeled test data, which is often unavailable in dynamic, real-world settings. Static, offline-trained models struggle to generalize, leading to frequent false positives.

To confront these challenges head-on, we introduce ALFred (Active Learning Framework for Real-world video anomaly detection Deployment), a novel approach tailored for pose-based VAD. Unlike conventional methods that rely on static training or passive online updates, ALFred leverages active learning to selectively query the most informative data points for labeling, optimizing efficiency and precision. This is paired with a human-in-the-loop mechanism, where human feedback refines the system’s understanding of context-specific anomalies, supported by AI-generated pseudo-labels. To enhance decision-making, ALFred incorporates a dynamic threshold adaptation strategy, adjusting detection boundaries based on historical labeled context and feedback — critical for handling non-percentage outputs and ensuring responsiveness in dynamic settings.

To rigorously evaluate ALFred’s reliability in real-world scenarios, we introduce the Error Balance Index (EBI), a novel metric that quantifies the trade-off between False Positive Rate (FPR) and False Negative Rate (FNR). Unlike traditional evaluation metrics that often overlook the delicate balance between misclassification types, EBI provides a more robust measure of a system’s practical viability in dynamic environments. We assessed ALFred in a lab-based simulation designed to emulate real-world conditions, including continual data and domain shifts. Building on established pose-based models suitable for online learning, such as STG-NF \cite{hirschorn2023stgnf}, TSGAD \cite{noghre2024tsgad}, and SPARTA \cite{noghre2024posewatch}. ALFred achieved an EBI of 68.91 for Q3, outperforming other methodologies. This success highlights its ability to deliver fast, privacy-preserving, and adaptable anomaly detection. Beyond performance, ALFred integrates contextual adaptability into real-world evaluation, paving the way for more reliable, deployment-ready VAD frameworks. Future improvements such as optimizing inference speed and expanding dataset diversity promise to further enhance its practical utility.

In summary, the contributions of this paper are: (1) Introducing ALFred, an active learning framework, applicable to existing models, for real-world VAD with adaptive thresholds; (2) Bringing human-in-the-loop functionality to introduce reliable training and validation sets for active learning; (3) Conducting extensive experiments that showcase the benefits of ALFred in adapting to unseen data.

\section{Related Works}
\label{sec:related}

Historically, traditional VAD methods relied on handcrafted features \cite{cheng2015gaussian, cocsar2016toward, yuan2014online, Senadeera_2024_CVPR, Khan_2024_CVPR}. These techniques were effective in controlled settings but struggled with generalization to real-world applications due to their reliance on predefined feature sets. The advent of deep learning revolutionized VAD, enabling the development of two primary approaches: pixel-based and pose-based. Pixel-based techniques \cite{huang2023multi, zaheer2022generative, georgescu2021background, georgescu2021anomaly, barbalau2023ssmtl++, wang2022video, chen2021nm, RL00, Gaus_2023vand, Lappas_2024vand7} analyze raw video data to detect irregularities, but their dependence on visual content raises privacy concerns and makes them susceptible to to environmental variations such as lighting changes and occlusions. In contrast, pose-based approaches extract skeletal movement data, offering a privacy-preserving alternative that is robust to background noise.

Semi-supervised learning strategies for pose-based VAD typically train models to internalize normal behavior by solving self-supervised tasks. These include reconstructing the current frame \cite{yu2023regularity, chen2023multiscale, jain2021posecvae, markovitz2020gepc, Heckler_2023vand3, Baradaran_2023vand4} or predicting past or future motion sequences \cite{zeng2021hierarchical, huang2022hierarchical, rodrigues2020multi}. Several studies \cite{noghre2024exploratory, morais2019learning} have adopted multi-branch frameworks that integrate both objectives, demonstrating anomaly detection performance.

Among publicly available pose-based models, GEPC, STG-NF, SPARTA and TSGAD have emerged as prominent baselines. GEPC \cite{markovitz2020gepc} employs a graph-based clustering approach where pose sequences are mapped into a latent space, with anomalies detected based on deviations from the learned distribution. STG-NF \cite{hirschorn2023stgnf} uses normalizing flows to transform input sequences into a standard distribution, flagging deviations as anomalies. TSGAD \cite{noghre2024tsgad} extends pose-based detection by incorporating trajectory analysis through a graph variational autoencoder, where discrepancies between predicted and actual trajectories contribute to the anomaly score. SPARTA \cite{noghre2025humancentricvideoanomalydetection, noghre2024posewatch} leverages the power of transformers for both reconstruction and prediction in order to teach the model to learn normal data. 

Despite the promise of these models, a major limitation is their reliance on static training paradigms. While some integrate online processing, they lack mechanisms for continuous adaptation to evolving behaviors in dynamic environments. Additionally, these models generally assume fixed thresholds for anomaly detection, which may not generalize well across different contexts. Addressing these issues requires incorporating active learning and human feedback to refine anomaly classification dynamically.

The concept of 'Online Anomaly Detection' varies across studies \cite{Doshi_2020_CVPR_Workshops, doshi2021online, doshi2023towards}. Many pixel-based approaches \cite{zhang2023online, karim2024real, Yang_2024vand5} define it as the ability to make real-time decisions without retraining, whereas our work focuses on continuous online learning, where models incrementally update their parameters using streaming data. Recent work \cite{Yao_2024_CVPR} underscores the challenges of deploying VAD models in dynamic environments, highlighting the need for adaptive methods that can handle evolving behaviors and environmental shifts. This distinction is crucial, as real-world anomaly detection systems must learn and adapt continuously, rather than relying on fixed decision boundaries set during initial training. Several studies \cite{doshi2023towards, danesh2023chad, Yang_2023vand2} have explored cross-domain evaluation, where models are tested on unseen datasets without prior exposure. However, these works do not propose explicit domain adaptation strategies, particularly in streaming settings where data distribution shifts over time. Effective deployment of VAD systems demands real-time domain adaptation mechanisms that can recalibrate models in response to new environmental conditions.

Overall, existing methods lack adaptive selection of informative samples and efficient human-in-the-loop refinement. Our proposed ALFred framework bridges this gap by integrating active learning with AI-generated pseudo-labels and dynamic thresholding, enabling continuous adaptation and reducing reliance on extensive manual annotations.
\section{Methodology}
\label{sec:methodology}

\subsection{ALFred}
\label{sec:alfred}

\begin{figure}[t]
  \centering
  \includegraphics[width=0.85\textwidth,,trim= 22 18 20 12,clip]{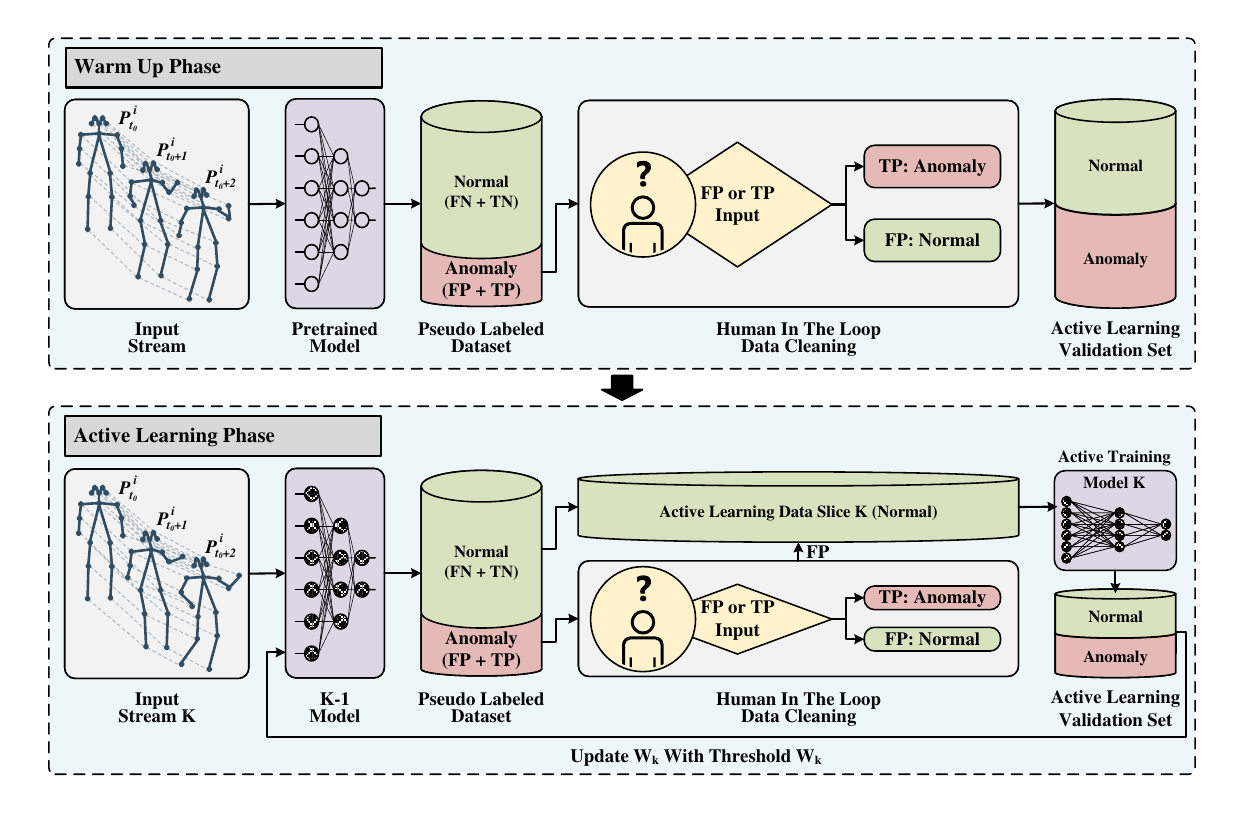}
  \caption{ALFred is a two-phase active learning framework for anomaly detection. The Warm-Up Phase utilizes a pretrained model to generate pseudo-labeled data, which is refined through human-in-the-loop validation before forming the active learning validation set. The Active Learning Phase iteratively improves the model by incorporating human feedback on false positives and true positives, updating the model threshold, and performing active training to enhance anomaly detection accuracy.}
  \label{fig:self_trans}
\end{figure}

Deploying VAD models in the real world requires more than high performance on standard benchmarks, it demands adaptability to unseen domains and precise threshold calibration, including frame preprocessing. ALFred addresses this challenge through a two-phase strategy: (1) a warm-up phase that collects a representative validation set from the target domain, and (2) an active learning phase that fine-tunes the model and threshold iteratively. The details are provided in Figure \ref{fig:self_trans}.

In pose-based VAD, advanced algorithms such as STG-NF \cite{hirschorn2023stgnf}, GEPC \cite{markovitz2020gepc}, and TSGAD \cite{noghre2024tsgad} are frequently evaluated using metrics like AUC-ROC and AUC-PR. While these metrics offer a robust evaluation of model performance across all possible thresholds, they do not directly provide a specific threshold suitable for real-world deployment. In practical settings, selecting an operational threshold is critical, as it governs the trade-off between false positive rate (FPR) and false negative rate (FNR), which varies depending on application-specific requirements. Additionally, the anomaly scores produced by VAD models are often not calibrated probabilities, complicating their direct interpretation as confidence measures.

To address threshold selection, equal error rate (EER) is defined as the point where FPR equals FNR, with more details in Section \ref{sec:metrics}. The EER is both model-specific and dependent on the training data. In a new environment, domain shifts can alter the underlying distribution of normal and anomalous events, rendering the original EER threshold ineffective. Thus, adapting the model and its threshold to the target domain is important in real-world scenarios. To establish an EER threshold specific to the target domain, labeled data from that domain is essential. This labeled data can serve as a validation set for the online-trained model, enabling the computation of an EER threshold tailored to the model's specific weights after training. To address this need efficiently, we introduce ALFred, an active learning framework designed to enhance the deployment of VAD models in real-world applications, consisting of two phases, including a warm-up phase that collects the validation set from the target domain and an active learning phase that deploys a training strategy for the algorithm.

\textbf{Warm Up Phase:} As illustrated in Figure \ref{fig:self_trans}, the warm-up phase of ALFred leverages a pre-trained model, initially calibrated with an EER threshold derived from the source domain, and deploys it in the target domain to assign pseudo-labels to the input skeleton data. These pseudo-labels classify segments of the data as normal or anomalous according to the threshold of the source domain. To address potential domain shift, human annotators review the pseudo-labeled anomalous data, correcting misclassifications and ensuring the labels accurately reflect the characteristics of the target domain. This human-in-the-loop approach results in a well-labeled active learning validation set, consisting of a balanced 50:50 ratio of normal to anomalous samples. This validation set is then utilized in the subsequent active learning phase to facilitate the model’s adaptation to the new environment.

\textbf{Active Learning Phase:}
In the active learning phase, the framework adapts the model to the target domain using the validation set curated during the warm-up phase. The pseudo-labeled normal data from the incoming stream serves as the training set \(K\), leveraging the real-world prevalence of normal instances over anomalies. Although some data points may be misclassified as FNs, they primarily act as noise, which can help the model learn more robust representations of normal behavior. Pseudo-labeled anomalies, on the other hand, are reviewed by human annotators, whose role is critical in identifying TPs and FPs. This human-in-the-loop process ensures high-quality feedback—especially for FPs, which are often challenging for automated systems to detect due to their ambiguous characteristics. These confirmed FPs are then added back to the training set K with the pseudo-labeled normal data, thereby improving the model’s understanding of normality and reducing future misclassifications.

It is important to note that the accuracy of pseudo-labels during this phase varies depending on the characteristics and performance of each model, and is tightly linked to the adaptive EER threshold established after each training iteration. This threshold, derived from the active learning validation set, is recalibrated iteratively to reflect the evolving decision boundary of the model, which is shaped by both the updated weights and the data characteristics in each slice \(K\). As a result, both the model and its EER threshold continuously adapt over time, facilitating stable and progressive learning, as illustrated in Figure \ref{fig:self_trans}.

\subsection{Base Algorithms}
As discussed in Section~\ref{sec:related}, pixel-based algorithms raise significant privacy concerns and face deployment challenges in real-world testbeds. To mitigate these issues and improve robustness, our study adopts a pose-based methodology using three recent state-of-the-art models with available code: STG-NF~\cite{hirschorn2023stgnf}, TSGAD~\cite{noghre2024tsgad}, and SPARTA~\cite{noghre2024posewatch}. TSGAD, originally a two-stream model, combines a Graph-Attentive Variational Autoencoder (GA-VAE) for pose reconstruction with trajectory prediction. For fair benchmarking, we use only the pose branch, training it via ELBO maximization to build a latent distribution of normal skeletal poses and detect anomalies based on distributional deviations. The graph-based design improves spatial relationship modeling between keypoints, enhancing anomaly localization in joint dynamics.

In contrast, STG-NF uses normalizing flows to map skeletal motion directly to a Gaussian distribution, providing exact likelihood scores for anomaly detection without extra heuristics. Its small size (1K parameters) makes it ideal for real-time and resource-limited environments. For consistency in evaluation, we reverse the original class labeling, treating anomalies as positive.

SPARTA~\cite{noghre2025humancentricvideoanomalydetection, noghre2024posewatch}, a transformer-based model, introduces spatio-temporal and relative pose tokenization with a shared encoder and dual decoders. The current decoder reconstructs present tokens, while the future decoder predicts upcoming ones. For diversity in model paradigms, we use only the prediction branch, enabling a comprehensive evaluation across three major self-supervised anomaly detection frameworks: reconstruction-based, probability-based, and prediction-based.

\subsection{Dataset \& Metrics}\label{sec:metrics}

For our framework evaluation, we utilize the HuVAD dataset \cite{pheva, pazho2025adaptivehumancentricvideoanomaly}, which contains over five million annotated frames collected from seven CCTV cameras across five days in a real-world community space. The dataset provides continuous recordings with human pose sequences and tracking IDs, supporting accurate cross-camera motion tracking for systematic evaluation of pose-based anomaly detection. It features a wide range of anomalous actions, including throwing, raising hands, lying down, falling, punching, kicking, pushing, and strangling, allowing for a comprehensive performance assessment in complex conditions. In particular, HuVAD includes a context-specific camera (CSC) that captures environments where typically anomalous actions, such as simulated fights, are considered normal due to security personnel training. This unique aspect enables evaluation of model adaptability and domain-specific learning, particularly useful for active learning-based anomaly detection frameworks.

Evaluation of anomaly detection models typically relies on metrics such as \textbf{AUC-ROC}, \textbf{AUC-PR}, and \textbf{EER}, each offering different advantages. AUC-ROC and AUC-PR evaluate performance across thresholds, with AUC-PR being particularly robust to class imbalance—critical in anomaly detection contexts. However, both metrics lack a concrete decision threshold, limiting their operational applicability. In contrast, \textbf{EER} (Equal Error Rate) provides a meaningful threshold where false positive and false negative rates are equal, making it valuable for deployment scenarios. To address limitations in existing metrics, we propose the \textbf{Error Balance Index (EBI)}, a novel metric that emphasizes the balance between FPR and FNR. While traditional metrics like Balanced Error Rate (BER) use the arithmetic mean of FPR and FNR, they inadequately capture the dominance of one error type over the other. In real-world settings, maintaining this balance is vital, as both false alarms and missed detections carry significant consequences. Drawing inspiration from the F1-score \cite{rijsbergen1979information}, EBI is defined as the harmonic mean of (1-FPR) and (1-FNR): $EBI=(2\times(1-FPR)\times (1-FNR))/((1-FPR) + (1-FNR))$, offering increased sensitivity to error imbalances and prioritizing misclassifications over correct predictions.


\section{Experiments and Evaluation}
\label{sec:experiments}
As detailed in our dataset description, the HuVAD dataset is specifically engineered to facilitate continual anomaly detection training. For each camera, the dataset is partitioned into two distinct subsets: a continual anomaly training set and a continual anomaly test set. The training set is further divided into nine slices to emulate a continual training environment. Analysis of pose counts for each continual set indicates that the anomaly rate in the training set remains below one percent per camera, whereas the test set exhibits an anomaly rate of approximately fifty percent, as detailed in Table \ref{tab:status}. To assess the performance of our proposed method, ALFred, we employ the continual test set from HuVAD as a simulated validation set. This setup mirrors data collected during an initial warm-up phase, reflecting real-world operational conditions. The nine slices of the continual training set are subsequently utilized during the active learning phase to refine ALFred’s capabilities.

\begin{table}[h]
\centering
\scriptsize
\begin{tabular}{c|rrc|rrc}
\toprule 
 & \multicolumn{3}{c|}{Active Learning Train Set} & \multicolumn{3}{c}{Active Learning Test Set} \\ \midrule
Cam ID & Normal & Anomaly & Anomaly Rate (\%) & Normal & Anomaly & Anomaly Rate (\%) \\ \midrule
0 & 430,659 & 4,216 & 0.97 & 26,052 & 26,093 & 49.96 \\
1 & 722,160 & 5,014 & 0.69 & 28,523 & 28,597 & 49.94 \\
2 & 661,280 & 6,813 & 1.02 & 25,402 & 25,449 & 49.95 \\
3 & 1,608,498 & 8,042 & 0.50 & 15,786 & 15,818 & 49.95 \\
4 & 415,838 & 1,530 & 0.37 & 37,208 & 37,274 & 49.96 \\
5 & 639,233 & 4,278 & 0.66 & 28,268 & 28,353 & 49.92 \\
CSC & 730,386 & 3,108 & 0.42 & 28,301 & 28,343 & 49.96 \\ \bottomrule 
\end{tabular}%
\caption{Total pose count and ratio for active learning from HuVAD\cite{pheva}}

\label{tab:status}
\end{table}

For initializing the weights of the source domain, referred to as K-1, we used the ShanghaiTech dataset \cite{shanghaitech}, a widely recognized benchmark in the VAD domain. These K-1 weights are first validated in the simulated validation set, namely, the HuVAD continual test set, to establish an EER threshold, customized to the target domain. This threshold is determined with human-in-the-loop guidance, based on the trade-off between FNR and FPR. Following this validation, the K-1 weights are used to assign pseudo-labels to incoming target domain data. The accuracy of these pseudo-labels is inherently dependent on the characteristics of each model and the defined EER threshold, and can vary during training. The final pseudo-label accuracy is evaluated against ground-truth labels from each data slice, using the overall FNR and FPR as reference metrics.

\textbf{Continual Learning} \cite{Yao_2024_CVPR}: No filtering is applied, and the training set includes all actual positives and negatives from the incoming data.

\textbf{Pseudo-Continual Learning}: Only data pseudo-labeled as normal by the K-1 weights is incorporated for training.

\textbf{Active Learning}: Human annotators identify FPs among the pseudolabeled anomalies - data incorrectly classified as anomalous but normal - and these FPs are added to the pseudolabeled normal data for training.

\textbf{AL Light}: A variant of active learning that minimizes annotator workload by sending only labeled positives with scores between the EER threshold and the rolling 50th percentile of recent max anomaly scores to human-in-the-loop review.

\begin{figure}[t]
  \centering
  \begin{minipage}[t]{0.48\textwidth}
    \centering
    \includegraphics[width=\textwidth, trim=0 0 10 0, clip]{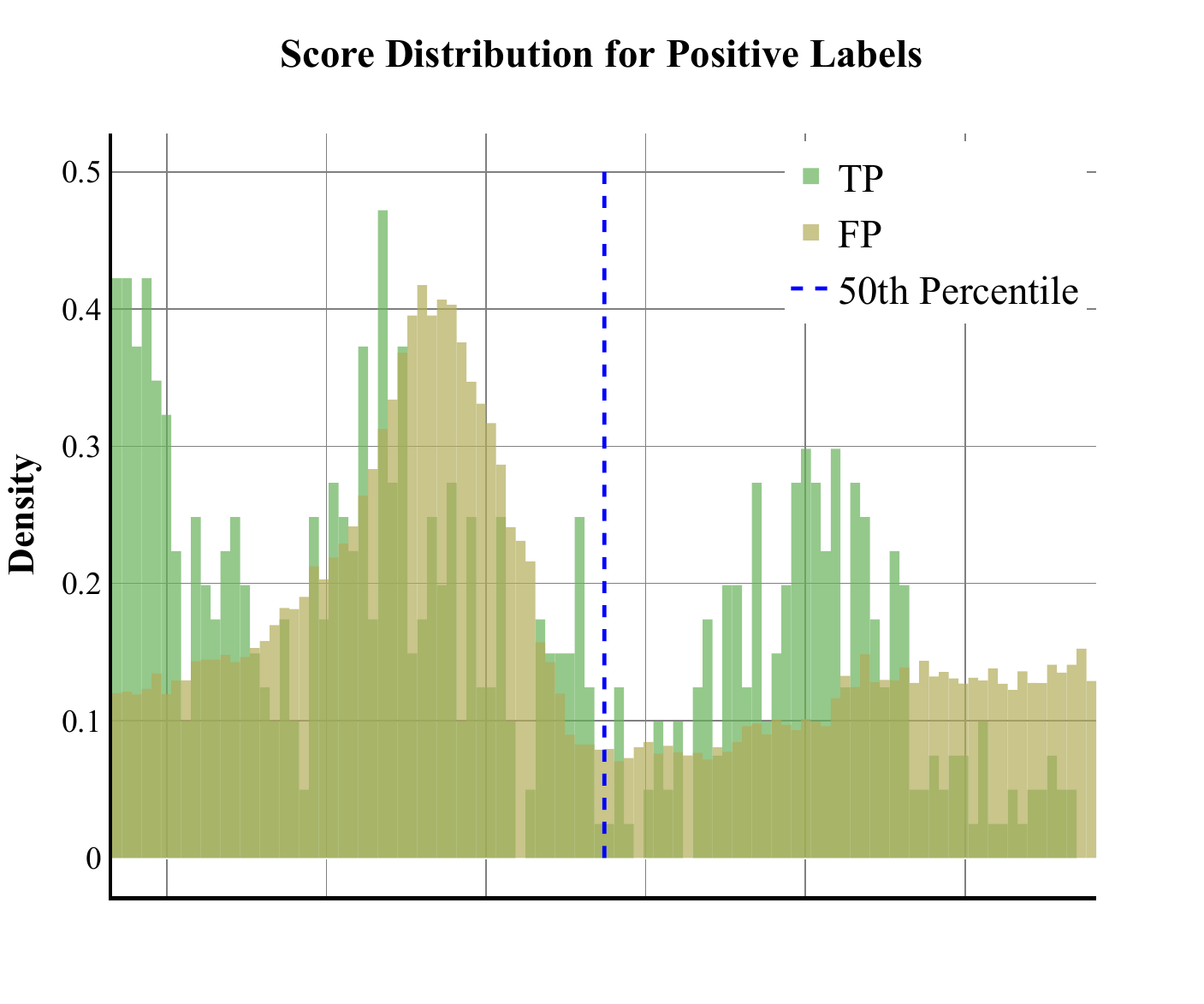}
    \caption{Score distribution for positive labels from C0 during Active Learning phase.}
    \label{fig:dist}
  \end{minipage}%
  \hfill
  \begin{minipage}[t]{0.48\textwidth}
    \centering
    \includegraphics[width=\textwidth, trim=0 0 10 0, clip]{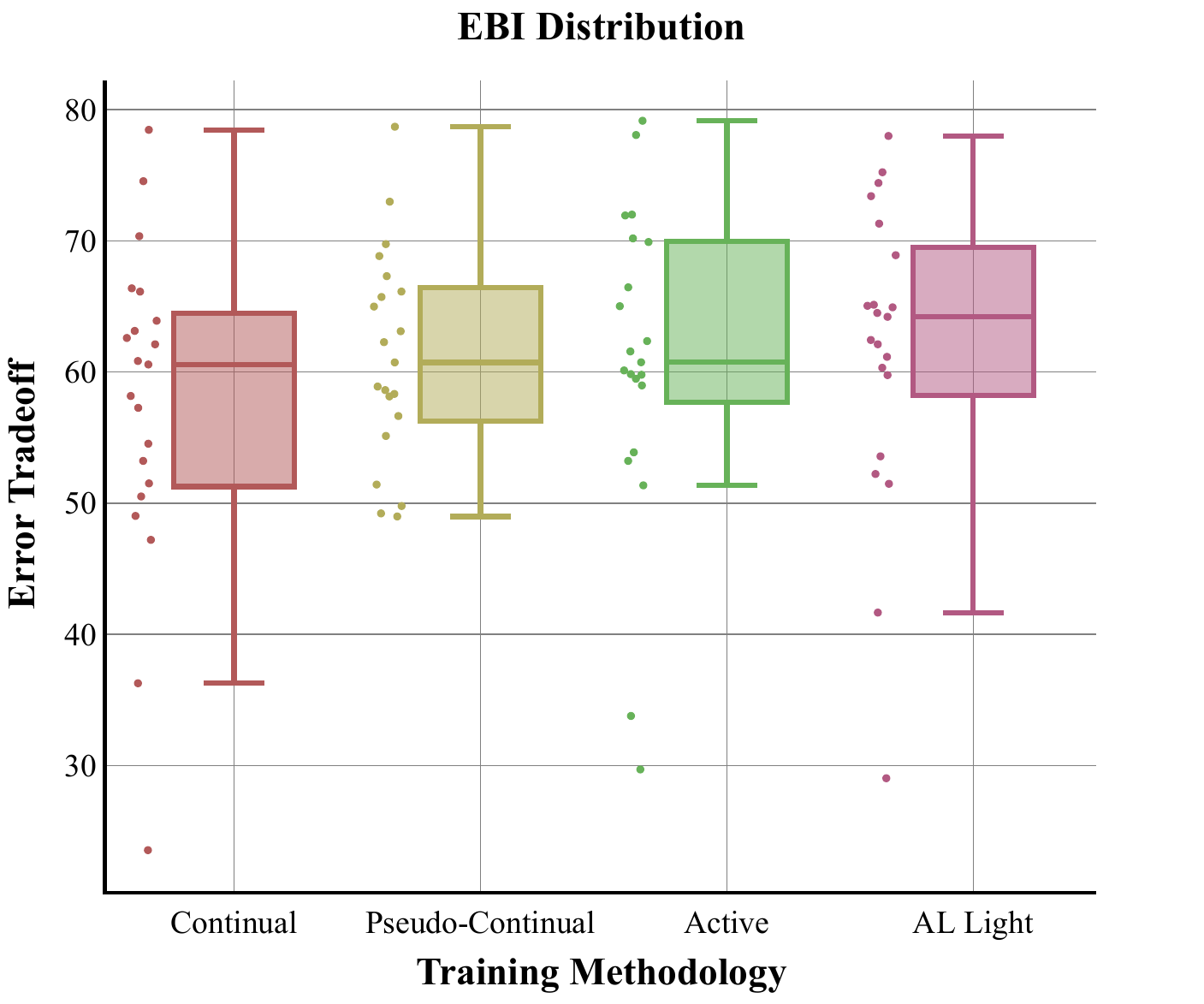}
    \caption{Error tradeoff distribution across different training methodologies.}
    \label{fig:ebi}
  \end{minipage}
\end{figure}

When discussing human-in-the-loop learning, our evaluation goes a step further by analyzing how to reduce annotator workload in an active learning scenario. Figure. \ref{fig:dist} illustrates the cumulative score distribution for positive labels during the active learning phase C0, using TSGAD \cite{noghre2024tsgad} as an example. The blue line represents the 50th percentile between the threshold and the maximum anomaly score at each step. Based on this analysis, limiting annotation requests to scores between the threshold and the midpoint can reduce the labeling workload by approximately 34.22\%. Due to the incremental nature of data feeding during training, we adopt a rolling 50th percentile approach, which consistently reduces the annotation load by at least 24.35\% at each step.

\begin{table}[]
\centering
\footnotesize
\begin{tabular}{cc|cc}
\toprule
\multicolumn{1}{c|}{Model} & Training Methodology & AUC-ROC & AUC-PR \\ \midrule
 &  & \multicolumn{2}{c}{CSC} \\ \midrule
\multicolumn{1}{c|}{\multirow{4}{*}{STG-NF}} & Offline Training & 43.71 & 48.29 \\
\multicolumn{1}{c|}{} & Continual Learning \cite{Yao_2024_CVPR} & 43.51 & 49.25 \\
\multicolumn{1}{c|}{} & Pseudo-Continual Learning & \textbf{44.51} & \textbf{49.51} \\
\multicolumn{1}{c|}{} & Active Learning & 43.61 & 49.21 \\ 
\multicolumn{1}{c|}{} & AL Light & 43.81 & 49.10 \\ \midrule
\multicolumn{1}{c|}{\multirow{4}{*}{TSGAD}} & Offline Training & 44.31 & 54.23 \\
\multicolumn{1}{c|}{} & Continual Learning \cite{Yao_2024_CVPR} & 47.36 & 55.69 \\
\multicolumn{1}{c|}{} & Pseudo-Continual Learning & 47.36 & 55.52 \\
\multicolumn{1}{c|}{} & Active Learning & \textbf{49.48} & \textbf{57.95} \\ 
\multicolumn{1}{c|}{} & AL Light & 48.13 & 57.25 \\ \midrule
\multicolumn{1}{c|}{\multirow{4}{*}{SPARTA}} & Offline Training & 45.24 & 50.36 \\
\multicolumn{1}{c|}{} & Continual Learning \cite{Yao_2024_CVPR} & 48.14 & 52.61 \\
\multicolumn{1}{c|}{} & Pseudo-Continual Learning & 48.11 & 52.60 \\
\multicolumn{1}{c|}{} & Active Learning & \textbf{48.48} & \textbf{52.86} \\ 
\multicolumn{1}{c|}{} & AL Light & 48.14 & 52.44 \\ \bottomrule 
\end{tabular}%

\caption{Comparison AUC-ROC and AUC-PR of CSC for models in different training methods.}
\label{tab:rocprcsc}
\end{table}

We initially assessed the performance of our models using the widely adopted metrics of AUC-ROC and AUC-PR. These metrics were applied to evaluate three distinct training data scenarios, with a baseline model established by training offline on the entire active learning training set without any data slicing. As shown in Table \ref{tab:rocprcsc}, using CSC as an example, all model weights were tested on the active learning validation set. Given the multi-step nature of the continual training process, we recorded the best results from each scenario to provide a comprehensive analysis. More details can be found in the supplementary materials for replication and further investigation.


While the findings align with recent work \cite{Yao_2024_CVPR}, confirming the effectiveness of our approach in achieving robust detection performance across diverse training conditions and highlighting its potential for practical VAD applications, the trends observed in Table \ref{tab:rocprcsc} do not necessarily translate to real-world scenarios. This discrepancy, primarily discussed in Section \ref{sec:alfred}, arises from the fact that these metrics are computed over varying thresholds and, by their nature, do not provide direct real-world insights. 

Since AUC-ROC and AUC-PR do not directly translate to operational performance in real-world scenarios, we adopted a different evaluation strategy for our continual training setup. Specifically, for each incoming data slice, we used the current model's EER threshold to assign pseudo-labels and computed TP, FP, TN, and FN by comparing these pseudo-labels to the ground truth. This method ensures a fair comparison across all scenarios by applying consistent pseudo-labeling criteria and is particularly appropriate for continual training, where offline training results cannot serve as a valid baseline due to the dynamic nature of the data stream. Such a comparison would be unfair given the static nature of offline training datasets. The accumulated results from each step for all cameras were used to calculate the overall FNR and FPR, as presented in Table \ref{tab:rocprcsc}. Detailed results for different cameras can be found in the supplementary materials for replication and further investigation.


\begin{table}[]
\centering
\scriptsize
\begin{tabular}{c|ccc|ccc|ccc}
\toprule
Training Methodology & FNR \% & FPR \% & EBI \% & FNR \% & FPR \% & EBI \% & FNR \% & FPR \% & EBI \% \\ \midrule
 & \multicolumn{3}{c|}{STG-NF \cite{hirschorn2023stgnf}} & \multicolumn{3}{c|}{TSGAD \cite{noghre2024tsgad}} & \multicolumn{3}{c}{SPARTA \cite{noghre2024posewatch}} \\ \midrule
Continual Learning \cite{Yao_2024_CVPR}& {\ul{32.77}} & 45.92 & 59.94 & {\ul{38.68}} & 38.20 & 61.56 & \textbf{24.25} & 41.55 & 65.98 \\
Pseudo-Continual & 37.74 & \textbf{28.01} & \textbf{66.77} & \textbf{29.84} & 47.82 & 59.85 & 25.67 & 40.61 & 66.02 \\
Active Learning & \textbf{35.24} & 39.85 & 62.36 & 41.29 & \textbf{31.29} & {\ul{63.31}} & {\ul{24.96}} & \textbf{39.46} & \textbf{67.01} \\
AL Light & 37.67 & {\ul{37.50}} & {\ul{62.56}} & 39.33 & {\ul{31.73}} & \textbf{64.24} & 24.86 & {\ul{39.78}} & {\ul{66.85}} \\
\bottomrule 
\end{tabular}
\caption{Comparison FNR and FPR of cumulative data for models in different training}
\label{tab:resultcombine}
\end{table}
Notably, while trends in AUC-ROC and AUC-PR on the validation set, do not always directly reflect model performance in real-world scenarios, these metrics remain essential within the ALFred framework for assessing model performance and determining the EER threshold. This threshold balances the trade-off between false positives and false negatives, which is critical for practical deployment. According to the study \cite{ardabili2024fpr}, FPR is particularly vital in real-world applications as it directly affects user trust and system reliability.

Based on Table \ref{tab:resultcombine}, the two selected models exhibit distinct performance patterns in real-world scenarios. The lighter model, STG-NF, demonstrates a lower FPR when trained using the pseudo-continual learning approach, which involves training on a smaller amount of data. However, as we transition to a larger and more complex dataset acquired through active learning, STG-NF fails to maintain this advantage, whereas TSGAD and SPARTA benefit from active learning, achieving improved FPR.


While the tables provide numerical insights, they do not explicitly illustrate the trade-off between different methodologies in terms of balancing FPR and FNR. To address this, we conducted an EBI analysis, introduced in Section \ref{sec:metrics}, which quantifies this balance and offers a clearer comparative assessment. As shown in Figure \ref{fig:ebi}, the EBI analysis reveals that Active Learning Light outperforms other methods, achieving the Q3 (25th percentile) at 68.91 and the highest median at 64.21. This indicates that, on average, Active Learning Light provides the most effective trade-off between reducing FPR and FNR, leading to more efficient anomaly classification. Compared to Continual Learning and Pseudo-Continual Learning, the higher Q3 and median suggest that Active Learning consistently achieves better overall reliability in most scenarios. The AL-Light method achieves comparable results to Active Learning while additionally reducing the human-in-the-loop workload by at least 24.35\%.
However, its wider range of values implies that performance variability is higher in extreme cases, indicating potential fluctuations under certain conditions. This trade-off underscores that while AL Light is the most effective approach for minimizing errors, careful parameter tuning and ongoing monitoring are necessary to manage its occasional instability.

\section{Research Questions and Future Directions}
\label{sec:future}

ALFred represents a first step toward adaptability in pose-based anomaly detection, but there remains room for improvement. One of the most critical areas for advancement is enhancing the accuracy of the framework, which can be achieved through multiple approaches. These include refining the model architecture, incorporating more diverse data, and leveraging advanced optimization techniques. One particularly promising method is reinforcement learning (RL), which enables active exploration of uncertain samples, allowing the model to prioritize ambiguous or low-likelihood instances for more effective learning. Additionally, adaptive anomaly score calibration through RL facilitates dynamic threshold optimization, improving robustness across varying conditions, while meta-learning and reward-based adaptation enhance generalization to unseen environments.

Beyond accuracy, another direction is evaluating the practical feasibility of the framework by assessing timing constraints, computational efficiency, and hardware requirements. Finally, achieving real-world applicability requires long-term deployment in operational settings, where the system can be continuously monitored to assess robustness, adaptability, and failure points. A structured evaluation and refinement cycle would provide critical insights into the system’s strengths and limitations, bridging the gap between research and real-world implementation. 
\section{Conclusion}
\label{sec:conclusion}


This paper addresses the challenges of deploying VAD models in real-world settings where domain shifts degrade performance. Conventional metrics like AUC-ROC and AUC-PR fail to provide actionable thresholds due to uncalibrated anomaly scores. To address this, we introduce ALFred, an active learning framework that efficiently adapts VAD models with minimal labeled data. ALFred employs a two-phase strategy: a warm-up phase using pseudo-labels refined by human annotators, and an active learning phase that fine-tunes the model with human feedback to establish an EER threshold. Experiments on the HuVAD dataset show that ALFred reduces false positives and enhances adaptability, demonstrating the potential of active learning for real-world VAD deployment. This work lays the foundation for further research into domain adaptation and human-in-the-loop refinement of anomaly detection systems.
\section*{Acknowledgment}
This research is supported by the National Science Foundation (NSF) under Award No. 1831795.
\bibliography{egbib}

\begin{thebibliography}{45}
\providecommand{\natexlab}[1]{#1}
\providecommand{\url}[1]{\texttt{#1}}
\expandafter\ifx\csname urlstyle\endcsname\relax
  \providecommand{\doi}[1]{doi: #1}\else
  \providecommand{\doi}{doi: \begingroup \urlstyle{rm}\Url}\fi

\bibitem[Alinezhad~Noghre et~al.(2023)Alinezhad~Noghre, Danesh~Pazho, Katariya, and Tabkhi]{RL00}
Ghazal Alinezhad~Noghre, Armin Danesh~Pazho, Vinit Katariya, and Hamed Tabkhi.
\newblock Understanding the challenges and opportunities of pose-based anomaly detection.
\newblock In \emph{Proceedings of the 8th International Workshop on Sensor-Based Activity Recognition and Artificial Intelligence}, iWOAR '23, New York, NY, USA, 2023. Association for Computing Machinery.
\newblock ISBN 9798400708169.
\newblock \doi{10.1145/3615834.3615844}.
\newblock URL \url{https://doi.org/10.1145/3615834.3615844}.

\bibitem[Ardabili et~al.(2024)Ardabili, Pazho, Noghre, Katariya, Hull, Reid, and Tabkhi]{ardabili2024fpr}
Babak~Rahimi Ardabili, Armin~Danesh Pazho, Ghazal~Alinezhad Noghre, Vinit Katariya, Gordon Hull, Shannon Reid, and Hamed Tabkhi.
\newblock Exploring public's perception of safety and video surveillance technology: A survey approach.
\newblock \emph{Technology in Society}, 78:\penalty0 102641, 2024.

\bibitem[Bao et~al.(2024)Bao, Sun, Deng, He, Zhang, and Li]{Bao_2024vand6}
Jinan Bao, Hanshi Sun, Hanqiu Deng, Yinsheng He, Zhaoxiang Zhang, and Xingyu Li.
\newblock Bmad: Benchmarks for medical anomaly detection.
\newblock In \emph{Proceedings of the IEEE/CVF Conference on Computer Vision and Pattern Recognition (CVPR) Workshops}, pages 4042--4053, June 2024.

\bibitem[Baradaran and Bergevin(2023)]{Baradaran_2023vand4}
Mohammad Baradaran and Robert Bergevin.
\newblock Multi-task learning based video anomaly detection with attention.
\newblock In \emph{Proceedings of the IEEE/CVF Conference on Computer Vision and Pattern Recognition (CVPR) Workshops}, pages 2886--2896, June 2023.

\bibitem[Barbalau et~al.(2023)Barbalau, Ionescu, Georgescu, Dueholm, Ramachandra, Nasrollahi, Khan, Moeslund, and Shah]{barbalau2023ssmtl++}
Antonio Barbalau, Radu~Tudor Ionescu, Mariana-Iuliana Georgescu, Jacob Dueholm, Bharathkumar Ramachandra, Kamal Nasrollahi, Fahad~Shahbaz Khan, Thomas~B Moeslund, and Mubarak Shah.
\newblock Ssmtl++: Revisiting self-supervised multi-task learning for video anomaly detection.
\newblock \emph{Computer Vision and Image Understanding}, 229:\penalty0 103656, 2023.

\bibitem[Chen et~al.(2021)Chen, Yue, Chang, Xu, and Jia]{chen2021nm}
Dongyue Chen, Lingyi Yue, Xingya Chang, Ming Xu, and Tong Jia.
\newblock Nm-gan: Noise-modulated generative adversarial network for video anomaly detection.
\newblock \emph{Pattern Recognition}, 116:\penalty0 107969, 2021.

\bibitem[Chen et~al.(2023)Chen, Kan, Zhang, Cen, Zhang, and Zhang]{chen2023multiscale}
Xiaoyu Chen, Shichao Kan, Fanghui Zhang, Yigang Cen, Linna Zhang, and Damin Zhang.
\newblock Multiscale spatial temporal attention graph convolution network for skeleton-based anomaly behavior detection.
\newblock \emph{Journal of Visual Communication and Image Representation}, 90:\penalty0 103707, 2023.

\bibitem[Cheng et~al.(2015)Cheng, Chen, and Fang]{cheng2015gaussian}
Kai-Wen Cheng, Yie-Tarng Chen, and Wen-Hsien Fang.
\newblock Gaussian process regression-based video anomaly detection and localization with hierarchical feature representation.
\newblock \emph{IEEE Transactions on Image Processing}, 24\penalty0 (12):\penalty0 5288--5301, 2015.

\bibitem[Co{\c{s}}ar et~al.(2016)Co{\c{s}}ar, Donatiello, Bogorny, Garate, Alvares, and Br{\'e}mond]{cocsar2016toward}
Serhan Co{\c{s}}ar, Giuseppe Donatiello, Vania Bogorny, Carolina Garate, Luis~Otavio Alvares, and Fran{\c{c}}ois Br{\'e}mond.
\newblock Toward abnormal trajectory and event detection in video surveillance.
\newblock \emph{IEEE Transactions on Circuits and Systems for Video Technology}, 27\penalty0 (3):\penalty0 683--695, 2016.

\bibitem[Danesh~Pazho et~al.(2023)Danesh~Pazho, Alinezhad~Noghre, Rahimi~Ardabili, Neff, and Tabkhi]{danesh2023chad}
Armin Danesh~Pazho, Ghazal Alinezhad~Noghre, Babak Rahimi~Ardabili, Christopher Neff, and Hamed Tabkhi.
\newblock Chad: Charlotte anomaly dataset.
\newblock In \emph{Scandinavian Conference on Image Analysis}, pages 50--66. Springer, 2023.

\bibitem[Doshi and Yilmaz(2020)]{Doshi_2020_CVPR_Workshops}
Keval Doshi and Yasin Yilmaz.
\newblock Continual learning for anomaly detection in surveillance videos.
\newblock In \emph{Proceedings of the IEEE/CVF Conference on Computer Vision and Pattern Recognition (CVPR) Workshops}, June 2020.

\bibitem[Doshi and Yilmaz(2021)]{doshi2021online}
Keval Doshi and Yasin Yilmaz.
\newblock Online anomaly detection in surveillance videos with asymptotic bound on false alarm rate.
\newblock \emph{Pattern Recognition}, 114:\penalty0 107865, 2021.

\bibitem[Doshi and Yilmaz(2023)]{doshi2023towards}
Keval Doshi and Yasin Yilmaz.
\newblock Towards interpretable video anomaly detection.
\newblock In \emph{Proceedings of the IEEE/CVF Winter Conference on Applications of Computer Vision}, pages 2655--2664, 2023.

\bibitem[Gaus et~al.(2023)Gaus, Bhowmik, Isaac-Medina, Shum, Atapour-Abarghouei, and Breckon]{Gaus_2023vand}
Yona Falinie~A. Gaus, Neelanjan Bhowmik, Brian K.~S. Isaac-Medina, Hubert P.~H. Shum, Amir Atapour-Abarghouei, and Toby~P. Breckon.
\newblock Region-based appearance and flow characteristics for anomaly detection in infrared surveillance imagery.
\newblock In \emph{Proceedings of the IEEE/CVF Conference on Computer Vision and Pattern Recognition (CVPR) Workshops}, pages 2995--3005, June 2023.

\bibitem[Georgescu et~al.(2021{\natexlab{a}})Georgescu, Barbalau, Ionescu, Khan, Popescu, and Shah]{georgescu2021anomaly}
Mariana-Iuliana Georgescu, Antonio Barbalau, Radu~Tudor Ionescu, Fahad~Shahbaz Khan, Marius Popescu, and Mubarak Shah.
\newblock Anomaly detection in video via self-supervised and multi-task learning.
\newblock In \emph{Proceedings of the IEEE/CVF conference on computer vision and pattern recognition}, pages 12742--12752, 2021{\natexlab{a}}.

\bibitem[Georgescu et~al.(2021{\natexlab{b}})Georgescu, Ionescu, Khan, Popescu, and Shah]{georgescu2021background}
Mariana~Iuliana Georgescu, Radu~Tudor Ionescu, Fahad~Shahbaz Khan, Marius Popescu, and Mubarak Shah.
\newblock A background-agnostic framework with adversarial training for abnormal event detection in video.
\newblock \emph{IEEE transactions on pattern analysis and machine intelligence}, 44\penalty0 (9):\penalty0 4505--4523, 2021{\natexlab{b}}.

\bibitem[Heckler et~al.(2023)Heckler, K\"onig, and Bergmann]{Heckler_2023vand3}
Lars Heckler, Rebecca K\"onig, and Paul Bergmann.
\newblock Exploring the importance of pretrained feature extractors for unsupervised anomaly detection and localization.
\newblock In \emph{Proceedings of the IEEE/CVF Conference on Computer Vision and Pattern Recognition (CVPR) Workshops}, pages 2917--2926, June 2023.

\bibitem[Hirschorn and Avidan(2023)]{hirschorn2023stgnf}
Or~Hirschorn and Shai Avidan.
\newblock Normalizing flows for human pose anomaly detection.
\newblock In \emph{Proceedings of the IEEE/CVF International Conference on Computer Vision}, pages 13545--13554, 2023.

\bibitem[Huang et~al.(2022)Huang, Liu, Zhang, Liu, Wen, Xu, and Wang]{huang2022hierarchical}
Chao Huang, Yabo Liu, Zheng Zhang, Chengliang Liu, Jie Wen, Yong Xu, and Yaowei Wang.
\newblock Hierarchical graph embedded pose regularity learning via spatio-temporal transformer for abnormal behavior detection.
\newblock In \emph{Proceedings of the 30th ACM International Conference on Multimedia}, pages 307--315, 2022.

\bibitem[Huang et~al.(2023)Huang, Zhao, Yu, Gao, and Wu]{huang2023multi}
Xiangyu Huang, Caidan Zhao, Jinhui Yu, Chenxing Gao, and Zhiqiang Wu.
\newblock Multi-level memory-augmented appearance-motion correspondence framework for video anomaly detection.
\newblock In \emph{2023 IEEE International Conference on Multimedia and Expo (ICME)}, pages 2717--2722. IEEE, 2023.

\bibitem[Jain et~al.(2021)Jain, Sharma, Velmurugan, and Banerjee]{jain2021posecvae}
Yashswi Jain, Ashvini~Kumar Sharma, Rajbabu Velmurugan, and Biplab Banerjee.
\newblock Posecvae: Anomalous human activity detection.
\newblock In \emph{2020 25th International Conference on Pattern Recognition (ICPR)}, pages 2927--2934. IEEE, 2021.

\bibitem[Karim et~al.(2024)Karim, Doshi, and Yilmaz]{karim2024real}
Hamza Karim, Keval Doshi, and Yasin Yilmaz.
\newblock Real-time weakly supervised video anomaly detection.
\newblock In \emph{Proceedings of the IEEE/CVF Winter Conference on Applications of Computer Vision}, pages 6848--6856, 2024.

\bibitem[Khan et~al.(2024)Khan, Ahmad, El~Saddik, Gueaieb, De~Masi, and Karray]{Khan_2024_CVPR}
Mustaqeem Khan, Jamil Ahmad, Abdulmotaleb El~Saddik, Wail Gueaieb, Giulia De~Masi, and Fakhri Karray.
\newblock Drone-hat: Hybrid attention transformer for complex action recognition in drone surveillance videos.
\newblock In \emph{Proceedings of the IEEE/CVF Conference on Computer Vision and Pattern Recognition (CVPR) Workshops}, pages 4713--4722, June 2024.

\bibitem[Lappas et~al.(2024)Lappas, Argyriou, and Makris]{Lappas_2024vand7}
Demetris Lappas, Vasileios Argyriou, and Dimitrios Makris.
\newblock Dynamic distinction learning: Adaptive pseudo anomalies for video anomaly detection.
\newblock In \emph{Proceedings of the IEEE/CVF Conference on Computer Vision and Pattern Recognition (CVPR) Workshops}, pages 3961--3970, June 2024.

\bibitem[Liu et~al.(2018)Liu, Luo, Lian, and Gao]{shanghaitech}
Wen Liu, Weixin Luo, Dongze Lian, and Shenghua Gao.
\newblock Future frame prediction for anomaly detection--a new baseline.
\newblock In \emph{Proceedings of the IEEE conference on computer vision and pattern recognition}, pages 6536--6545, 2018.

\bibitem[Markovitz et~al.(2020)Markovitz, Sharir, Friedman, Zelnik-Manor, and Avidan]{markovitz2020gepc}
Amir Markovitz, Gilad Sharir, Itamar Friedman, Lihi Zelnik-Manor, and Shai Avidan.
\newblock Graph embedded pose clustering for anomaly detection.
\newblock In \emph{Proceedings of the IEEE/CVF Conference on Computer Vision and Pattern Recognition}, pages 10539--10547, 2020.

\bibitem[Morais et~al.(2019)Morais, Le, Tran, Saha, Mansour, and Venkatesh]{morais2019learning}
Romero Morais, Vuong Le, Truyen Tran, Budhaditya Saha, Moussa Mansour, and Svetha Venkatesh.
\newblock Learning regularity in skeleton trajectories for anomaly detection in videos.
\newblock In \emph{Proceedings of the IEEE/CVF conference on computer vision and pattern recognition}, pages 11996--12004, 2019.

\bibitem[Noghre et~al.(2024{\natexlab{a}})Noghre, Pazho, and Tabkhi]{noghre2024exploratory}
Ghazal~Alinezhad Noghre, Armin~Danesh Pazho, and Hamed Tabkhi.
\newblock An exploratory study on human-centric video anomaly detection through variational autoencoders and trajectory prediction.
\newblock In \emph{Proceedings of the IEEE/CVF Winter Conference on Applications of Computer Vision}, pages 995--1004, 2024{\natexlab{a}}.

\bibitem[Noghre et~al.(2024{\natexlab{b}})Noghre, Pazho, and Tabkhi]{noghre2024posewatch}
Ghazal~Alinezhad Noghre, Armin~Danesh Pazho, and Hamed Tabkhi.
\newblock Posewatch: A transformer-based architecture for human-centric video anomaly detection using spatio-temporal pose tokenization.
\newblock \emph{arXiv preprint arXiv:2408.15185}, 2024{\natexlab{b}}.

\bibitem[Noghre et~al.(2024{\natexlab{c}})Noghre, Pazho, and Tabkhi]{noghre2024tsgad}
Ghazal~Alinezhad Noghre, Armin~Danesh Pazho, and Hamed Tabkhi.
\newblock An exploratory study on human-centric video anomaly detection through variational autoencoders and trajectory prediction.
\newblock In \emph{Proceedings of the IEEE/CVF Winter Conference on Applications of Computer Vision}, pages 995--1004, 2024{\natexlab{c}}.

\bibitem[Noghre et~al.(2024{\natexlab{d}})Noghre, Yao, Pazho, Ardabili, Katariya, and Tabkhi]{pheva}
Ghazal~Alinezhad Noghre, Shanle Yao, Armin~Danesh Pazho, Babak~Rahimi Ardabili, Vinit Katariya, and Hamed Tabkhi.
\newblock Pheva: A privacy-preserving human-centric video anomaly detection dataset.
\newblock \emph{arXiv preprint arXiv:2408.14329}, 2024{\natexlab{d}}.

\bibitem[Noghre et~al.(2025)Noghre, Pazho, and Tabkhi]{noghre2025humancentricvideoanomalydetection}
Ghazal~Alinezhad Noghre, Armin~Danesh Pazho, and Hamed Tabkhi.
\newblock Human-centric video anomaly detection through spatio-temporal pose tokenization and transformer, 2025.
\newblock URL \url{https://arxiv.org/abs/2408.15185}.

\bibitem[Pazho et~al.(2025)Pazho, Yao, Noghre, Ardabili, Katariya, and Tabkhi]{pazho2025adaptivehumancentricvideoanomaly}
Armin~Danesh Pazho, Shanle Yao, Ghazal~Alinezhad Noghre, Babak~Rahimi Ardabili, Vinit Katariya, and Hamed Tabkhi.
\newblock Towards adaptive human-centric video anomaly detection: A comprehensive framework and a new benchmark, 2025.
\newblock URL \url{https://arxiv.org/abs/2408.14329}.

\bibitem[Rijsbergen(1979)]{rijsbergen1979information}
Van Rijsbergen.
\newblock Information retrieval; ; butterworth, 1978.
\newblock \emph{J. librariansh.}, 11:\penalty0 237, 1979.

\bibitem[Rodrigues et~al.(2020)Rodrigues, Bhargava, Velmurugan, and Chaudhuri]{rodrigues2020multi}
Royston Rodrigues, Neha Bhargava, Rajbabu Velmurugan, and Subhasis Chaudhuri.
\newblock Multi-timescale trajectory prediction for abnormal human activity detection.
\newblock In \emph{Proceedings of the IEEE/CVF winter conference on applications of computer vision}, pages 2626--2634, 2020.

\bibitem[Senadeera et~al.(2024)Senadeera, Yang, Kollias, and Slabaugh]{Senadeera_2024_CVPR}
Damith~Chamalke Senadeera, Xiaoyun Yang, Dimitrios Kollias, and Gregory Slabaugh.
\newblock Cue-net: Violence detection video analytics with spatial cropping enhanced uniformerv2 and modified efficient additive attention.
\newblock In \emph{Proceedings of the IEEE/CVF Conference on Computer Vision and Pattern Recognition (CVPR) Workshops}, pages 4888--4897, June 2024.

\bibitem[Wang et~al.(2022)Wang, Wang, Qin, Zhang, Bao, and Huang]{wang2022video}
Guodong Wang, Yunhong Wang, Jie Qin, Dongming Zhang, Xiuguo Bao, and Di~Huang.
\newblock Video anomaly detection by solving decoupled spatio-temporal jigsaw puzzles.
\newblock In \emph{European Conference on Computer Vision}, pages 494--511. Springer, 2022.

\bibitem[Yang and Radke(2024)]{Yang_2024vand5}
Zhengye Yang and Richard~J. Radke.
\newblock Context-aware video anomaly detection in long-term datasets.
\newblock In \emph{Proceedings of the IEEE/CVF Conference on Computer Vision and Pattern Recognition (CVPR) Workshops}, pages 4002--4011, June 2024.

\bibitem[Yang et~al.(2023)Yang, Soltani, and Darve]{Yang_2023vand2}
Ziyi Yang, Iman Soltani, and Eric Darve.
\newblock Anomaly detection with domain adaptation.
\newblock In \emph{Proceedings of the IEEE/CVF Conference on Computer Vision and Pattern Recognition (CVPR) Workshops}, pages 2958--2967, June 2023.

\bibitem[Yao et~al.(2024)Yao, Noghre, Pazho, and Tabkhi]{Yao_2024_CVPR}
Shanle Yao, Ghazal~Alinezhad Noghre, Armin~Danesh Pazho, and Hamed Tabkhi.
\newblock Evaluating the effectiveness of video anomaly detection in the wild: Online learning and inference for real-world deployment.
\newblock In \emph{Proceedings of the IEEE/CVF Conference on Computer Vision and Pattern Recognition (CVPR) Workshops}, pages 4832--4841, June 2024.

\bibitem[Yu et~al.(2023)Yu, Zhao, Fang, Deng, Su, Wang, Gan, Lu, and Wu]{yu2023regularity}
Shoubin Yu, Zhongyin Zhao, Haoshu Fang, Andong Deng, Haisheng Su, Dongliang Wang, Weihao Gan, Cewu Lu, and Wei Wu.
\newblock Regularity learning via explicit distribution modeling for skeletal video anomaly detection.
\newblock \emph{IEEE Transactions on Circuits and Systems for Video Technology}, 2023.

\bibitem[Yuan et~al.(2014)Yuan, Fang, and Wang]{yuan2014online}
Yuan Yuan, Jianwu Fang, and Qi~Wang.
\newblock Online anomaly detection in crowd scenes via structure analysis.
\newblock \emph{IEEE transactions on cybernetics}, 45\penalty0 (3):\penalty0 548--561, 2014.

\bibitem[Zaheer et~al.(2022)Zaheer, Mahmood, Khan, Segu, Yu, and Lee]{zaheer2022generative}
M~Zaigham Zaheer, Arif Mahmood, M~Haris Khan, Mattia Segu, Fisher Yu, and Seung-Ik Lee.
\newblock Generative cooperative learning for unsupervised video anomaly detection.
\newblock In \emph{Proceedings of the IEEE/CVF conference on computer vision and pattern recognition}, pages 14744--14754, 2022.

\bibitem[Zeng et~al.(2021)Zeng, Jiang, Ding, Li, Hao, and Qiu]{zeng2021hierarchical}
Xianlin Zeng, Yalong Jiang, Wenrui Ding, Hongguang Li, Yafeng Hao, and Zifeng Qiu.
\newblock A hierarchical spatio-temporal graph convolutional neural network for anomaly detection in videos.
\newblock \emph{IEEE Transactions on Circuits and Systems for Video Technology}, 33\penalty0 (1):\penalty0 200--212, 2021.

\bibitem[Zhang et~al.(2023)Zhang, Song, Jiang, and Li]{zhang2023online}
Yuxing Zhang, Jinchen Song, Yuehan Jiang, and Hongjun Li.
\newblock Online video anomaly detection.
\newblock \emph{Sensors}, 23\penalty0 (17):\penalty0 7442, 2023.

\end{thebibliography}

\section{Benchmarking}

The following equations are applied during the Active Learning Light phase to reduce the workload for the human-in-the-loop process:

\begin{equation}
    \theta_{\text{med}}^{(t)} = \text{Percentile}(\{s_{t-N}, \dots, s_tmax\}, 50)
\end{equation}

\begin{equation}
    \text{Send to annotator if } \theta_{\text{EER}} \leq s_t \leq \theta_{\text{med}}^{(t)}
\end{equation}

\subsection{Conventional Benchmarking}
\label{sec:con}

Tables \ref{tab:roc-pr} and \ref{tab:rocprcsc} provide a more detailed view of how each model performs under different training methodologies. The best results across each K step are presented, and these findings align with study \cite{Yao_2024_CVPR}, confirming the effectiveness of our approach in achieving robust detection performance across diverse training conditions.

\begin{table*}[]
\centering
\tiny
\begin{tabular}{cc|cc|cc|cc}
\toprule\toprule
\multicolumn{1}{c|}{Model} & Training Methodology & AUC-ROC & AUC-PR & AUC-ROC & AUC-PR & AUC-ROC & AUC-PR \\ \midrule
 &  & \multicolumn{2}{c|}{C0} & \multicolumn{2}{c|}{C1} & \multicolumn{2}{c}{C2} \\ \midrule
\multicolumn{1}{c|}{\multirow{4}{*}{STG-NF}} & Offline Training & 58.64 & 57.78 & 50.08 & 47.73 & 61.82 & 56.03 \\
\multicolumn{1}{c|}{} & Continual Learning \cite{Yao_2024_CVPR} & 58.65 & 58.91 & 51.53 & 50.63 & \textbf{64.31} & 59.29 \\
\multicolumn{1}{c|}{} & Pseudo-Continual Learning & \textbf{59.91} & 58.82 & \textbf{53.65} & 51.19 & 62.43 & \textbf{59.69} \\
\multicolumn{1}{c|}{} & Active Learning & 59.64 & \textbf{59.12} & 53.36 & \textbf{51.63} & 63.54 & 59.32 \\ 
\multicolumn{1}{c|}{} & AL Light & 59.32 & 58.98 & 53.21 & 51.55 & 64.21 & 58.96 \\ \midrule
\multicolumn{1}{c|}{\multirow{4}{*}{TSGAD}} & Offline Training & 58.74 & 58.29 & 62.75 & 68.64 & 62.34 & 61.34 \\
\multicolumn{1}{c|}{} & Continual Learning \cite{Yao_2024_CVPR} & \textbf{59.60} & 59.95 & \textbf{65.28} & 69.33 & \textbf{67.86} & 63.36 \\
\multicolumn{1}{c|}{} & Pseudo-Continual Learning & 59.38 & 61.91 & 64.53 & \textbf{69.75} & 64.79 & \textbf{65.17} \\
\multicolumn{1}{c|}{} & Active Learning & 59.22 & \textbf{62.43} & 59.52 & 65.18 & 61.78 & 62.36 \\ 
\multicolumn{1}{c|}{} & AL Light & 59.27 & 62.01 & 61.32 & 67.91 & 63.51 & 63.75 \\  \midrule
\multicolumn{1}{c|}{\multirow{4}{*}{SPARTA}} & Offline Training & 58.95 & 58.29 & 54.17 & 50.57 & \textbf{63.86} & 57.45 \\
\multicolumn{1}{c|}{} & Continual Learning \cite{Yao_2024_CVPR} & 59.87 & 58.67 & 55.82 & 51.98 & 60.86 & 60.18 \\
\multicolumn{1}{c|}{} & Pseudo-Continual Learning & 59.93 & 58.72 & 55.90 & 52.01 & 61.09 & 60.29 \\
\multicolumn{1}{c|}{} & Active Learning & \textbf{59.93} & \textbf{58.72} & 56.48 & \textbf{52.40} & 61.99 & 61.00 \\ 
\multicolumn{1}{c|}{} & AL Light & 59.65 & 58.60 & \textbf{56.54} & 52.33 & 62.26 & \textbf{61.43} \\ \midrule
&  & \multicolumn{2}{c|}{C3} & \multicolumn{2}{c|}{C4} & \multicolumn{2}{c}{C5} \\ \midrule
\multicolumn{1}{c|}{\multirow{4}{*}{STG-NF}} & Offline Training & 52.18 & 50.87 & \textbf{61.46} & 56.41 & \textbf{61.08} & 57.63 \\
\multicolumn{1}{c|}{} & Continual Learning \cite{Yao_2024_CVPR} & \textbf{54.52} & \textbf{53.59} & 61.46 & 56.41 & 59.41 & 56.94 \\
\multicolumn{1}{c|}{} & Pseudo-Continual Learning & 50.90 & 50.23 & 61.32 & \textbf{58.87} & 57.92 & 56.28 \\
\multicolumn{1}{c|}{} & Active Learning & 53.07 & 52.38 & 61.13 & 56.26 & 59.65 & 57.10 \\ 
\multicolumn{1}{c|}{} & AL Light & 52.97 & 52.65 & 61.09 & 56.31 & 59.33 & \textbf{57.89} \\ \midrule
\multicolumn{1}{c|}{\multirow{4}{*}{TSGAD}} & Offline Training & 63.85 & 58.35 & 62.51 & 64.05 & 64.33 & 63.45 \\
\multicolumn{1}{c|}{} & Continual Learning \cite{Yao_2024_CVPR} & 67.41 & 66.72 & 62.51 & 64.05 & 70.15 & 67.38 \\
\multicolumn{1}{c|}{} & Pseudo-Continual Learning & \textbf{73.04} & \textbf{75.52} & \textbf{67.04} & \textbf{68.89} & \textbf{70.25} & 66.36 \\
\multicolumn{1}{c|}{} & Active Learning & 68.84 & 70.56 & 63.55 & 63.80 & 66.92 & \textbf{68.86} \\ 
\multicolumn{1}{c|}{} & AL Light & 69.96 & 72.32 & 63.12 & 64.50 & 68.23 & 67.79 \\ \midrule
\multicolumn{1}{c|}{\multirow{4}{*}{SPARTA}} & Offline Training & 56.68 & 53.95 & 61.38 & 55.80 & 60.22 & 55.88 \\
\multicolumn{1}{c|}{} & Continual Learning \cite{Yao_2024_CVPR} & 64.13 & 58.48 & 64.13 & \textbf{58.48} & 64.60 & 59.71 \\
\multicolumn{1}{c|}{} & Pseudo-Continual Learning & 64.96 & 59.37 & 64.16 & 58.27 & 64.72 & 59.77 \\
\multicolumn{1}{c|}{} & Active Learning & 65.07 & \textbf{59.50} & 64.32 & 58.42 & \textbf{64.93} & \textbf{59.90} \\
\multicolumn{1}{c|}{} & AL Light & \textbf{66.31} & 59.22 & \textbf{64.41} & 58.44 & 64.55 & 59.82 \\ \bottomrule \bottomrule
\end{tabular}
\caption{Comparison of AUC-ROC and AUC-PR for models in different training methods.}
\label{tab:roc-pr}
\end{table*}

\subsection{Real World Benchmarking}
\label{sec:real}

\begin{table*}[]
\centering
\footnotesize
\begin{tabular}{cc|cc|cc|cc}
\toprule\toprule
\multicolumn{1}{c|}{Model} & Training Methodology & FNR \% & FPR \% & FNR \% & FPR \% & FNR \% & FPR \% \\ \midrule
 &  & \multicolumn{2}{c|}{C0} & \multicolumn{2}{c|}{C1} & \multicolumn{2}{c}{C2} \\ \midrule
\multicolumn{1}{c|}{\multirow{3}{*}{STG-NF}} & Continual Learning & \textbf{34.23} & 33.00 & \ul{27.42} & 56.32 & \textbf{22.91} & 59.36 \\
\multicolumn{1}{c|}{} & Pseudo-Continual Learning & 48.46 & \textbf{21.34} & 33.55 & \textbf{28.59} & 35.27 & \textbf{32.40} \\
\multicolumn{1}{c|}{} & Active Learning & \ul{42.36} & 21.54 & \textbf{23.21} & 51.04 & 32.29 & \ul{37.47} \\ 
\multicolumn{1}{c|}{} & AL Light & 44.52 & \ul{21.39} & 31.05 & \ul{39.41} & \ul{29.77} & 39.28 \\ \midrule
\multicolumn{1}{c|}{\multirow{3}{*}{TSGAD}} & Continual Learning & 27.80 & 51.29 & 62.64 & 17.08 & \textbf{16.09} & 49.39 \\
\multicolumn{1}{c|}{} & Pseudo-Continual Learning & 26.00 & \textbf{42.06} & \textbf{29.28} & 38.62 & \ul{37.67} & 44.19 \\
\multicolumn{1}{c|}{} & Active Learning & \textbf{13.26} & 53.26 & 54.63 & \textbf{10.93} & 51.83 & \textbf{22.25} \\ 
\multicolumn{1}{c|}{} & AL Light & \ul{19.07} & \ul{46.78} & \ul{48.34} & \ul{12.60} & 46.12 & \ul{26.68} \\ \midrule
\multicolumn{1}{c|}{\multirow{3}{*}{SPARTA}} & Continual Learning & \textbf{27.49} & 47.59 & \textbf{21.60} & 54.89 & \ul{23.23} & \ul{47.16} \\
\multicolumn{1}{c|}{} & Pseudo-Continual Learning & 29.63 & \ul{46.58} & \ul{23.59} & 53.08 & 23.81 & \textbf{46.14} \\
\multicolumn{1}{c|}{} & Active Learning & 28.98 & \textbf{45.66} & 25.67 & \ul{51.12} & \textbf{21.93} & 48.09 \\ 
\multicolumn{1}{c|}{} & AL Light & \ul{28.32} & 46.67 & 23.97 & \textbf{50.00} & 22.37 & 47.78 \\ \midrule
 &  & \multicolumn{2}{c|}{C3} & \multicolumn{2}{c|}{C4} & \multicolumn{2}{c}{C5} \\ \midrule
\multicolumn{1}{c|}{\multirow{3}{*}{STG-NF}} & Continual Learning & 45.11 & 16.83 & 40.44 & 85.33 & \textbf{24.61} & 76.12 \\
\multicolumn{1}{c|}{} & Pseudo-Continual Learning & \textbf{37.42} & \textbf{12.46} & 42.02 & \textbf{40.72} & \ul{32.14} & \textbf{61.39} \\
\multicolumn{1}{c|}{} & Active Learning & \ul{40.96} & \ul{13.47} & \textbf{41.02} & \ul{80.16} & 35.37 & 77.14 \\ 
\multicolumn{1}{c|}{} & AL Light & 42.69 & 13.61 & \ul{41.70} & 80.68 & 38.94 & \ul{68.39} \\ \midrule
\multicolumn{1}{c|}{\multirow{3}{*}{TSGAD}} & Continual Learning & \ul{40.44} & 35.11 & 61.23 & \textbf{39.69} & 43.31 & \ul{34.97} \\
\multicolumn{1}{c|}{} & Pseudo-Continual Learning & \textbf{21.75} & 53.51 & 42.12 & 56.32 & \textbf{19.85} & 57.99 \\
\multicolumn{1}{c|}{} & Active Learning & 47.38 & \ul{30.65} & \ul{41.38} & 51.26 & \ul{30.43} & \textbf{25.52} \\ 
\multicolumn{1}{c|}{} & AL Light & 47.90 & \textbf{29.93} & \textbf{40.70} & \ul{51.15} & 29.34 & 28.02 \\ \midrule

\multicolumn{1}{c|}{\multirow{3}{*}{SPARTA}} & Continual Learning & 26.73 & 15.54 & \textbf{30.44} & 62.15 & \textbf{16.18} & 63.85 \\
\multicolumn{1}{c|}{} & Pseudo-Continual Learning & \ul{26.36} & 15.48 & 32.37 & 61.60 & 17.81 & 62.59 \\
\multicolumn{1}{c|}{} & Active Learning & \textbf{26.29} & \textbf{14.53} & 31.15 & \ul{59.05} & 17.55 & \textbf{59.99} \\
\multicolumn{1}{c|}{} & AL Light & 27.74 & \ul{15.27} & \ul{31.02} & \textbf{58.94} & \ul{17.06} & \ul{61.89} \\ \bottomrule \bottomrule
\end{tabular}
\caption{Comparison of FNR and FPR for models in different training method}
\label{tab:results}
\end{table*}

\begin{table}[]
\centering
\footnotesize
\begin{tabular}{cc|cc}
\toprule\toprule
\multicolumn{1}{c|}{Model} & Training Methodology & FNR \% & FPR \% \\ \midrule
 &  & \multicolumn{2}{c}{CSC} \\ \midrule
\multicolumn{1}{c|}{\multirow{3}{*}{STG-NF}} & Continual Learning & 32.55 & {\ul{16.68}} \\
\multicolumn{1}{c|}{} & Pseudo-Continual Learning & 41.57 & \textbf{13.47} \\
\multicolumn{1}{c|}{} & Active Learning & \textbf{26.14} & 17.20 \\ 
\multicolumn{1}{c|}{} & AL Light & {\ul{31.11}} & 17.13\\ \midrule
\multicolumn{1}{c|}{\multirow{3}{*}{TSGAD}} & Continual Learning & 26.47 & 43.49 \\
\multicolumn{1}{c|}{} & Pseudo-Continual Learning & 51.57 & \textbf{31.79} \\
\multicolumn{1}{c|}{} & Active Learning & {\ul{26.14}} & 33.63 \\ 
\multicolumn{1}{c|}{} & AL Light & \textbf{15.62} & {\ul{33.45}} \\ \midrule
\multicolumn{1}{c|}{\multirow{3}{*}{SPARTA}} & Continual Learning & {\ul{25.42}} & 33.41 \\
\multicolumn{1}{c|}{} & Pseudo-Continual Learning & 34.71 & 30.55 \\
\multicolumn{1}{c|}{} & Active Learning & 26.27 & {\ul{29.64}} \\ 
\multicolumn{1}{c|}{} & AL Light & \textbf{23.59} & \textbf{29.36} \\ \bottomrule \bottomrule
\end{tabular}
\caption{Comparison FNR and FPR of CSC for models in different
training method}
\label{tab:resultcsc}
\end{table}

Tables \ref{tab:results} and \ref{tab:resultcsc} provide a more detailed view of how each model performs under different training methodologies. The FNR and FPR values represent the cumulative counts of false negatives and false positives across all steps during training.

\end{document}